\pdfoutput=1
\documentclass[11pt]{article}

\usepackage[]{acl}

\usepackage{times}
\usepackage{latexsym}
\usepackage{float}
\usepackage{algorithm}
\usepackage{algpseudocode}

\usepackage[T1]{fontenc}
\usepackage[utf8]{inputenc}

\usepackage{microtype}
\usepackage{cleveref}
\usepackage{graphicx}
\usepackage{url}

\crefformat{section}{\S#2#1#3} % see manual of cleveref, section 8.2.1
\crefformat{subsection}{\S#2#1#3}
\crefformat{subsubsection}{\S#2#1#3}

\title{Adapting Transformer Language Models for Predictive Typing \\
in Brain-Computer Interfaces}

\author{Shijia Liu \and David A. Smith
\\ Khoury College of Computer Sciences \\ Northeastern University \\ Boston, MA \\ \texttt{liu.shij@northeastern.edu, dasmith@ccs.neu.edu}
 }

\begin{document}
\maketitle
\begin{abstract}

Brain-computer interfaces (BCI) are an important mode of alternative and augmentative communication for many people.  Unlike keyboards, many BCI systems do not display even the 26 letters of English at one time, let alone all the symbols in more complex systems.  Using language models to make character-level predictions, therefore, can greatly speed up BCI typing \citep{ghosh2017neural}. While most existing BCI systems employ character n-gram models or no LM at all, this paper adapts several wordpiece-level Transformer LMs to make character predictions and evaluates them on typing tasks. GPT-2 fares best on clean text, but different LMs react differently to noisy histories. We further analyze the effect of character positions in a word and context lengths.

\end{abstract}

\section{Introduction}
The use of state-of-the-art language models in facilitating typing tasks is prominent in various computer applications nowadays. From autocomplete features in email and text messaging apps to search suggestion functionalities in many websites, language models have been playing an important roles in reducing users' typing time and thus enhancing their overall experience by providing accurate typing predictions. 

One distinctive use of language models in predictive typing is in brain-computer interface (BCI) systems. BCI systems enable people who have lost their motor functions and cannot communicate via common writing and speaking channels to express themselves by brain waves \citep{birbaumer1999spelling, sellers2010brain}. Specifically, these systems observe signals, such as electroencephalograms (EEG), and process them in order to detect users' intended expressions. 

During typing in BCI systems, candidate letters are often presented to users in series, so it is desirable to show more likely target letters earlier to save typing time. Thus, language models become useful since they are able to make relatively accurate typing suggestions as well as correct errors \citep{ghosh2017neural}. Currently, studies on using language models to facilitate BCI typing have limited scope: the language models examined are mostly simple character n-gram models \citep{oken2014brain, mora2014language, speier2016integrating, speier2017online, speier2018improving} or other finite-state approaches \citep{dudy2018multi}. \citet{dong2019noisy} shows an LSTM based language model for typing prediction that is robust to noisy input, but it requires training on noisy data.

In the past few years, large-scale pretrained language models have proved to be best suited for many NLP tasks due to the significant performance improvements they are able to achieve. Hence, these models are good candidates to incorporate into BCI systems. In this scenario, we would like language models to give accurate predictions on what users intend to type. For such a task, autoregressive language models, which are trained to sequentially generate text, have shown to be very effective. This class of models---starting with GPT \citep{radford2018improving}, GPT-2 \citep{radford2019language}, and Transformer-XL \citep{dai-etal-2019-transformer}---generate coherent paragraphs of text that are increasingly hard to distinguish from human output.

For text generation tasks, these autoregressive language models are typically evaluated on metrics such as perplexity (PPL), bits-per-character (BPC) or word level accuracy. However, since in BCI systems candidate characters are presented sequentially to users, the rankings of the target letters matter the most. Thus, information retrieval measures such as mean reciprocal rank (MRR) and recall at rank $k$ should be the main evaluation metrics when judging predictive performance. It is essential to achieve high mean reciprocal rank on the correct letters and high recall of the correct characters in the top $k$ candidates presented to the users.

In this paper, therefore, we evaluate several current pretrained language models on a predictive typing task.  We chose models that are large enough to have proven effective on several tasks in prior work, but small enough to run on a single machine for deployment in a BCI system.  The language models we evaluate are all based on the transformer architecture \citep{vaswani2017attention}, which has become the standard choice for many NLP problems. We seek to predict the next characters the users intend to type given the input so far.  The research questions we aim to answer are: (1) What is the overall performance on character prediction of the different models that we evaluate? How does the performance differ when predicting different character positions in words? (2) How does the predictive performance vary with different context length (we model context length as the number of complete words in given input)? (3) In the presence of noise, how does the performance degradation diverge for different models, and what are the reasons for such difference? Further, is such degradation correlated with different context lengths? 

Four different transformer language models are examined in order to address these questions: Reformer \citep{Kitaev2020Reformer}, Transformer-XL \citep{dai-etal-2019-transformer}, GPT-2 \citep{radford2019language} and GPT \citep{radford2018improving}. We also utilize a unigram baseline model to compare results. %We realize that there are other transformer-based autoregressive language models that are also suitable for predictive typing tasks. However, as a proof of concept, we choose these four since they represent a valid variety of language models that have different properties. 
Two datasets we use to run experiments on are the ALS Phraseset \citep{costello2014message} and the Switchboard Corpus \citep{godfrey1992switchboard}.

Evaluation results indicate that GPT-2 gives best performance on both evaluation datasets across all metrics with clean input text. Predicting the first characters in words proves to be the most difficult for all transformer models, where it generally gets easier when predicting the later parts of words. Longer context length usually helps improve the predicting performance, although not always for predicting the first characters in words. Finally, different models react differently to input noise, where some of the best performing models on clean text degrade significantly with the presence of noise. With some variation, the performance degradation with noise is generally correlated with context lengths. 

\section{Related Work}

There have been multiple attempts to integrate language models into BCI systems. \citet{oken2014brain} utilizes a six-gram character level language model, where the difference between the predicted probabilities of the target letters and the top ranked letters is used to mark the levels of spelling difficulty. \citet{speier2016integrating, speier2017online, speier2018improving} is a series of work that investigates how typing rate and accuracy could be improved with the help of language models in BCI systems by carrying out actual user studies. \citet{dudy2018multi} proposes an FST based joint word-character language model that incorporates ambiguous history into decoding in order to build a more robust prediction system, which achieves promising results on perplexity and predictive accuracy. Lastly, \citet{dong2019noisy} examines the usefulness of training recurrent language models on noisy text since it makes the models more robust to typical noisy input in BCI systems.

\section{Models and Methods}

\subsection{Models} \label{subsec: models}

As stated above, we mainly consider pretrained transformer-based autoregressive language models for character prediction. The models we use for our experiments are Reformer, Transformer-XL, GPT-2 and GPT. We briefly review the main ideas of these models below:

\begin{itemize}
    \item \textbf{Reformer} \citep{Kitaev2020Reformer}: Reformer is an autoregressive transformer language model with techniques that allow improved computing efficiency. The techniques include replacing dot-product attention by LSH (locality-sensitive hashing) attention, as well as using reversible residual layers instead of standard residuals. The model we use for our experiments is a character level Reformer model trained on enwik8 \citep{hutter2018}.
    \item \textbf{Transformer-XL} \citep{dai-etal-2019-transformer}: Transformer-XL introduces a novel recurrence mechanism that allows the model to learn longer-term dependencies beyond a fixed length context. The segment-level recurrence mechanism (a segment is a number of consecutive tokens), which considers inputs from two consecutive segments, enables the model to capture information from the previous segment as well as the current one.
    \item \textbf{GPT-2} \citep{radford2019language}: GPT-2 is a 1.5B parameter transformer pretrained on a large and diverse English corpus. It is trained to predict the next words in sentences. The training process is autoregressive, meaning that the target sequences of the model are the same as the input sequences, shifted one token to the right. When making predictions, the model can only see input from history, not from future. Hence, the model is suitable for text generation tasks.
    \item \textbf{GPT} \citep{radford2018improving}: GPT is the previous version of GPT-2. It has the same autoregressive training objective and is trained on the BookCorpus dataset \citep{zhu2015aligning}. The model has 117M parameters.
\end{itemize}

\subsection{Methods}

In this section we detail the methods we use to generate character level predictions from what is returned by language models in \cref{subsec: models}.

\subsubsection{Reformer: Direct Character Level Predictions}

The Reformer model is a character level model, which returns a probabilistic distribution over a set of characters in its vocabulary for next character prediction. We simply use this distribution as final result.

\subsubsection{Transformer-XL: Whole Word Prediction with a Closed Vocabulary}
The Transformer-XL model utilizes a closed vocabulary of English words. We simply take the input and remove the partially-typed last word. The model gives a prediction of the last word over the entire vocabulary. Then we select words in the vocabulary that have prefixes matching the partially-typed last word from the input. Then we renormalize the probability distribution over the selected words and marginalize over the first character after the matched prefix. The resulting character-level distribution is a prediction of the next character given current input.

\subsubsection{GPT-2 and GPT: Beam Search on Subword Units}
As for GPT-2 and GPT models, byte-pair encoding (BPE) \citep{sennrich-etal-2016-neural} is used for tokenization. The main idea of BPE is to create a fixed-size vocabulary that contains common English subword units. A less common word would thus be broken down into several subword units in the vocabulary. For example, in GPT-2, the tokenization of character sequence \emph{``peanut butter and jel"} would be (``\_'' denotes a space character):
\begin{center}
    \emph{['pe', 'anut', '\_butter', '\_and', '\_j', 'el']}
\end{center}

Therefore, in order to generate a predictive distribution on the next characters, we need to examine all the possibilities that could complete the final subword units in the input sequences. Similar to what we do for the Transformer-XL model, we make the model predict the last subword unit over the entire subword vocabulary. Then we select the subword units with prefixes matching the partially-typed last subword from the input (in the example case above, the selected subword units should start with \emph{``el"}). We then perform the same renormalization and marginalization as for the Transformer-XL model and obtain a distribution over a set of characters. 

However, predicting only the final subword units may limit the space of possible continuations of the input. To remedy that, we need to look beyond one subword unit. We therefore use beam search to generate multiple subword units. Beam search is a heuristic search algorithm that expands the most promising hypotheses in a selected set. In our case, the most promising hypotheses would be the top ranked subword units on the current beam. For each of these hypotheses, we make the language model further predict the next subword unit over the entire vocabulary. Then the top ranked hypotheses returned are again retained for further expansion. The renormalization and marginalization are performed at the end of the beam search for all candidates on the beam. The beam size and the beam search depth are hyperparameters that can be adjusted.

\section{Data and Experiments}

\subsection{Data}

In our experiments we evaluate on two datasets. The first is closely related to the BCI domain: it contains manually anonymized messages created by people with ALS (\textbf{A}myotrophic \textbf{L}ateral \textbf{S}clerosis, a progressive nervous system disease that causes loss of muscle control). The second dataset is the Switchboard corpus \citep{godfrey1992switchboard}, which is a collection of transcripts of two-sided telephone conversations. We choose this dataset since we would like to simulate the conversational nature of communications between BCI users and systems. Below we will give more detailed information of these two datasets.

\begin{itemize}
    \item \textbf{ALS Phraseset} \citep{costello2014message}: The dataset contains 2,082 messages that serve as a ``vocabulary'' in a message bank. The creation of these message is a part of a process called \emph{voice banking}. Voice banking is a procedure of ``recording a large inventory of someone's speech which is then used to create a synthetic voice that approximates their natural voice'' \citep{costello2014message}. This enables people to communicate messages via a voice synthesizer that imitates their natural speech. The message vocabulary in this dataset can be retrieved by users of augmentative communication technologies and they can speak them in their own voice.
    \item \textbf{Switchboard Corpus} \citep{godfrey1992switchboard}: The Switchboard Corpus contains transcripts from about 260 hours of speech. Specifically, it includes about 2,400 two-sided telephone conversations among 543 speakers from around the United States. The corpus was collected automatically over T1 lines at Texas Instruments. The callers are given appropriate recorded prompts, and topics for discussion are also introduced before the conversations start. 
    
\end{itemize}

\subsection{Experiments}

In this section, we describe the experiments performed for our analysis. We first outline data preprocessing steps and introduce evaluation metrics. Then we inspect results from various experiments in detail.

\subsubsection{Data Preprocessing} \label{sec:preprocessing}

For the ALS phraseset, we remove all the punctuation marks and expand all contractions. As for the Switchboard corpus, we delete interjections such as ``uh-huh", ``yeah" and ``uh", as well as punctuation and dialogue-specific placeholders such as ``MUMBLEx''.

For evaluation, we use the whole ALS phraseset and randomly sample 500 conversations from Switchboard. For our experiments on character predictions, we only look at the first four turns of the Switchboard conversations.

\subsubsection{Evaluation Metrics}

The main evaluation metrics for our character prediction task is MRR and Recall@\emph{k}. We report and compare MRR@\emph{k} and Recall@\emph{k} across different models, where \emph{k} indicates the number of top-ranked characters in our predictions that will be taken into account when computing the metrics. Specifically, MRR@\emph{k} is the mean over different predictions of the reciprocal rank of the correct target letter if it is in the top \emph{k} characters or zero otherwise. Recall@\emph{k} indicates the proportion of the cases where the correct target letter is in the top \emph{k} characters suggested by the language model. 

\subsection{Main Results on Character Prediction} \label{sec:main_results_char_pred}

In this section, we report the main results of our character prediction task on the ALS Phraseset and the Switchboard Corpus. We examine results averaged over all character positions in words, as well as results for different character positions. For the ALS Phraseset, we let the language models predict the characters in the last word of each phrase. For the Switchboard Corpus, the last word of each turn is used for prediction. %Furthermore, we add a unigram language model trained on the ALS Phraseset as a baseline for comparison. 
Furthermore, we use a unigram language model trained on the ALS Phraseset to interpolate with all the transformer models (except for the character level model Reformer) in our experiments for smoothing purposes, since the distributions given by the transformer models can be very ``sharp", where most of the probability mass falls on the top few letters. Therefore, interpolating with a unigram model gives more probability to the lower ranked letters. We choose an interpolation coefficient of 0.8. As for the beam search method we use for GPT-2 and GPT, the beam size is chosen to be 20 where the search depth is 2. We use the Huggingface transformers v4.17.0 \footnote{\url{https://huggingface.co/docs/transformers/index}} implementations for all transformer language models and run our experiments with their default parameter settings. For all experiments, we report the mean results of 3 consecutive runs. We run our experiments on a computing cluster with GPUs and it takes $\sim$9 hours to complete one run on both datasets.

\begin{table*}
\centering
    \begin{tabular}{c|cccccc}
    \hline
    Language Model & MRR@10 & Recall@10 & MRR@5 & Recall@5 & MRR@3 & Recall@3 \\
    \hline
    Unigram Baseline & 0.2294 & 0.7022 & 0.1835 & 0.3681 & 0.1552 & 0.2424 \\
    Reformer & 0.7080 & 0.9381 & 0.6984 & 0.8664 & 0.6806 & 0.7883\\
    Transformer-XL & 0.7267 & 0.9308 & 0.7185 & 0.8700 & 0.7039 & 0.8060 \\
    GPT: Beam Search & 0.7776 & 0.9487 & 0.7706 & 0.8973 & 0.7607 & 0.8545 \\
    GPT-2: Beam Search & \textbf{0.8253} & \textbf{0.9611} & \textbf{0.8196} & \textbf{0.9195} & \textbf{0.8103} & \textbf{0.8789} \\
    \hline
    \end{tabular}
    \caption{Results on the ALS Phraseset, averaged across all character positions}
    \label{tab:als_phraseset_avg}
\end{table*}

\begin{table*}
\centering
    \begin{tabular}{c|cccccc}
    \hline
    Language Model & MRR@10 & Recall@10 & MRR@5 & Recall@5 & MRR@3 & Recall@3 \\
    \hline
    Unigram Baseline & 0.2292 & 0.7184 & 0.1821 & 0.3669 & 0.1543 & 0.2420 \\
    Reformer & 0.7361 & 0.9394 & 0.7265 & 0.8708 & 0.7115 & 0.8050 \\
    Transformer-XL & 0.7648 & 0.9419 & 0.7558 & 0.8756 & 0.7447 & 0.8270 \\
    GPT: Beam Search & 0.6474 & 0.8960 & 0.6316 & 0.7774 & 0.6180 & 0.7175 \\
    GPT-2: Beam Search & \textbf{0.8185} & \textbf{0.9530} & \textbf{0.8122} & \textbf{0.9055} & \textbf{0.8028} & \textbf{0.8650} \\
    \hline
    \end{tabular}
    \caption{Results on the Switchboard Corpus, averaged across all character positions}
    \label{tab:switchboard_avg}
\end{table*}

\paragraph{Averaging over all character positions in a word} Table~\ref{tab:als_phraseset_avg} shows the results on the ALS phraseset, averaged over all character positions. To give a straightforward sense of the good performance all transformer language models deliver, we include a unigram baseline model (trained on the ALS Phraseset). As we can see, all transformer language models perform significantly better than the unigram baseline model. Among all models, GPT-2 fares best across all metrics, achieving a 82.53\% MRR@10 and a 96.11\% Recall@10. %All transformer models perform significantly better than the unigram baseline. 
    Interestingly, the Reformer model, which is trained at the character level, does not perform as well as other transformer models trained at word (or subword) level. This is probably due to the fact that the corpus that the Reformer model is trained on (enwik8) is relatively smaller and less diverse compared to the training corpus of other transformer models. 
    
    Table~\ref{tab:switchboard_avg} shows the results on the Switchboard Corpus, averaged over all character positions. Same as what we see on the ALS phraseset, GPT-2 gives the best performance across all metrics, achieving a 81.85\% MRR@10 and a 95.30\% Recall@10. Interesting, GPT does not perform as well as it does on the ALS phraseset. In fact, it has the worst results among all models. This is probably because the BookCorpus dataset it is trained on \citep{zhu2015aligning} has little in common with conversational transcripts. The Reformer and the Transformer-XL models perform reasonably well. Similar to the what we have for the ALS phraseset, Reformer as a character based transformer model does not fare as well as word based models such as Transformer-XL and GPT-2.
    
\begin{figure}[h]
    \includegraphics[width=\linewidth]{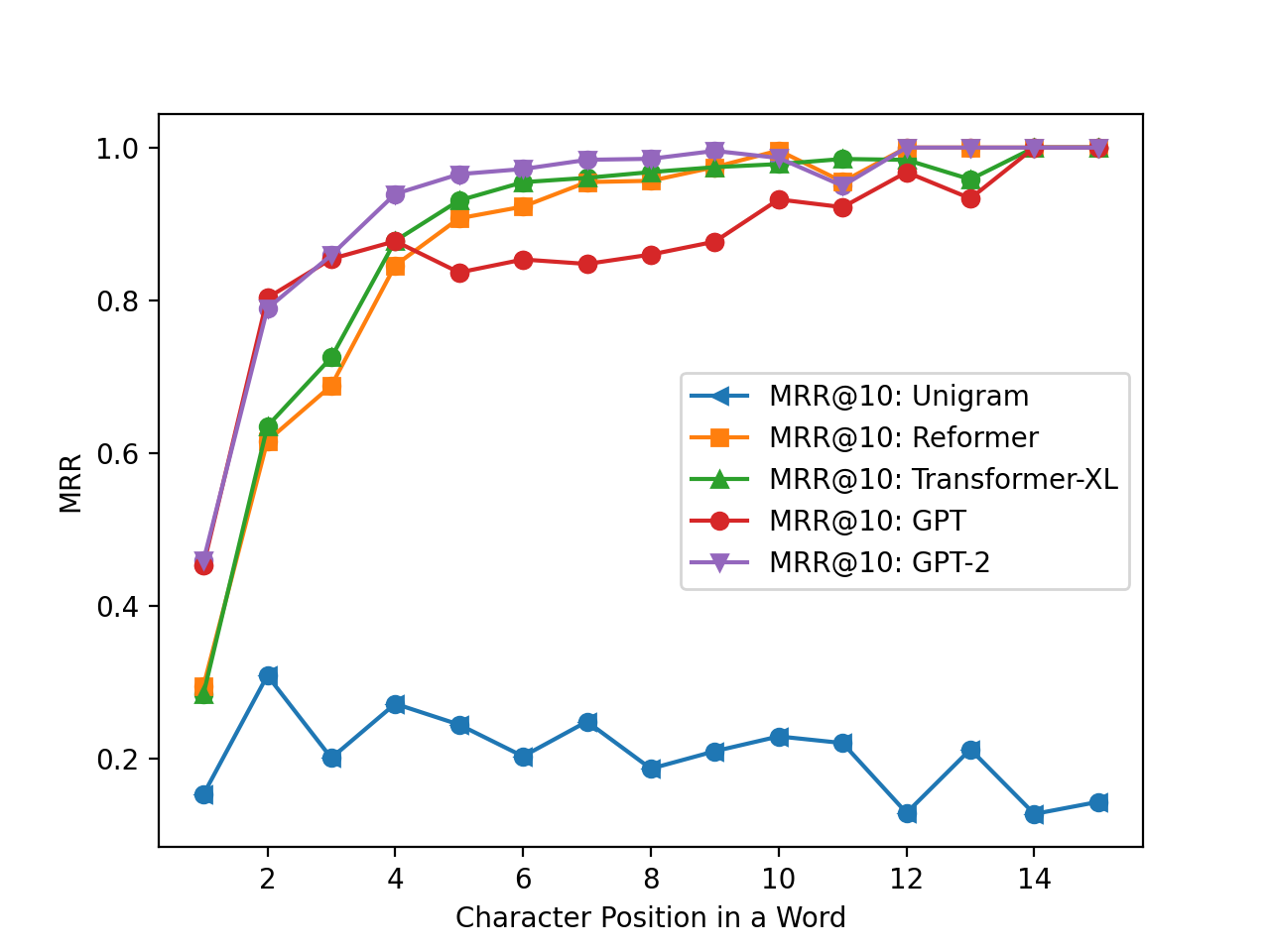}
    \caption{MRR@10 for different character positions on ALS Phraseset}
    \label{fig:mrr@10_als_char_pos}
\end{figure}

\begin{figure}[h]
    \includegraphics[width=\linewidth]{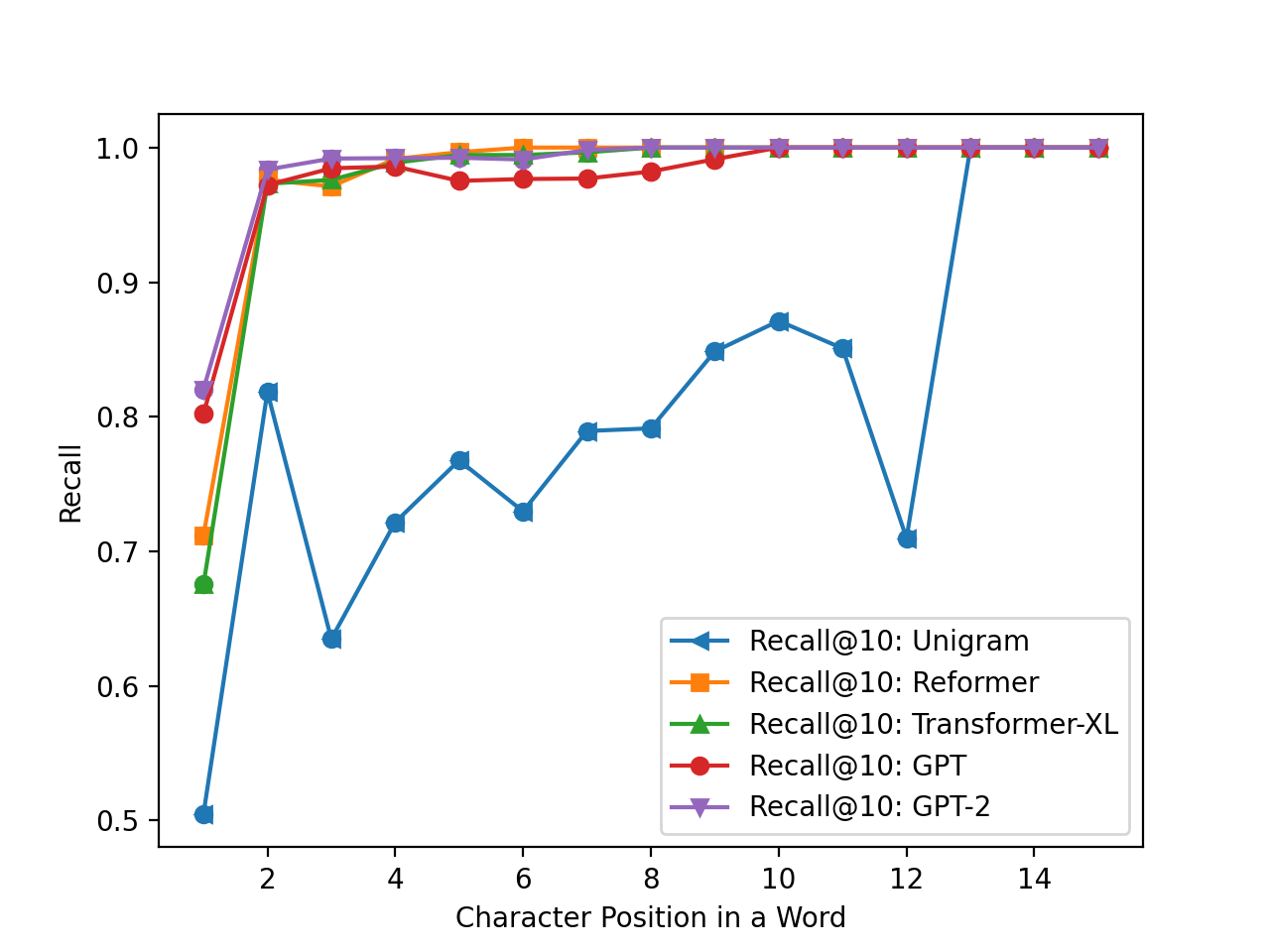}
    \caption{Recall@10 for different character positions on ALS Phraseset}
    \label{fig:recall@10_als_char_pos}
\end{figure}

\paragraph{On specific positions in a word} Figure~\ref{fig:mrr@10_als_char_pos} and Figure~\ref{fig:recall@10_als_char_pos} illustrate the results on MRR@10 and Recall@10 for different character positions in the predicted words on the ALS Phraseset. As shown in these figures, predicting the first character in a word proves to be the most difficult across all transformer models. GPT-2 and GPT fare better in predicting the first character than Reformer and Transformer-XL do. As we have more characters appearing in a word as context, the performance starts to improve for all transformer models, and the gap between models begins to decrease. All transformer models achieves near-perfect results in Recall@10 when we have more than 8 characters typed in a word.
    
\begin{figure}[h]
    \includegraphics[width=\linewidth]{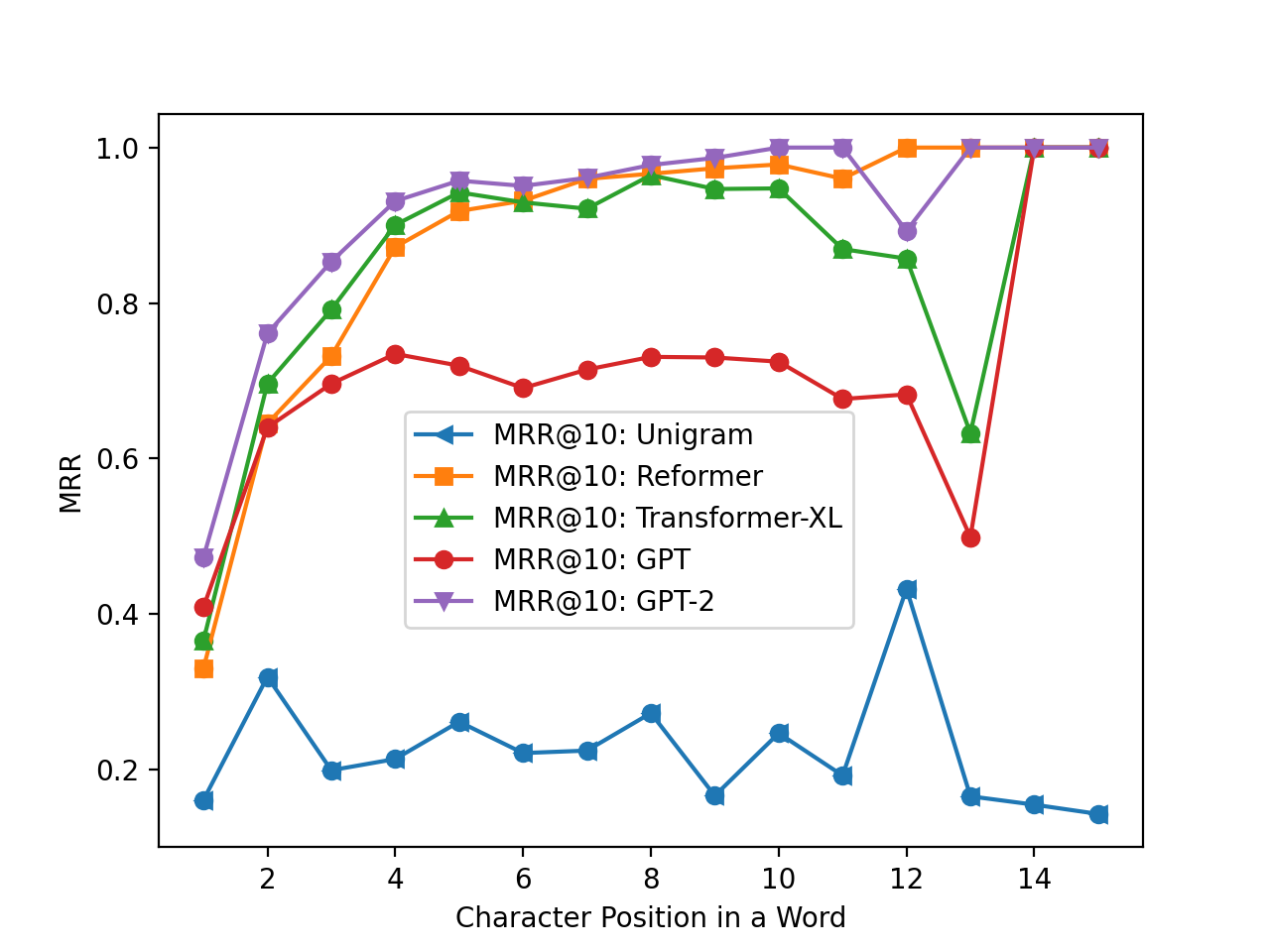}
    \caption{MRR@10 for different character positions on Switchboard Corpus}
    \label{fig:mrr@10_swbd_char_pos}
\end{figure}

\begin{figure}[h]
    \includegraphics[width=\linewidth]{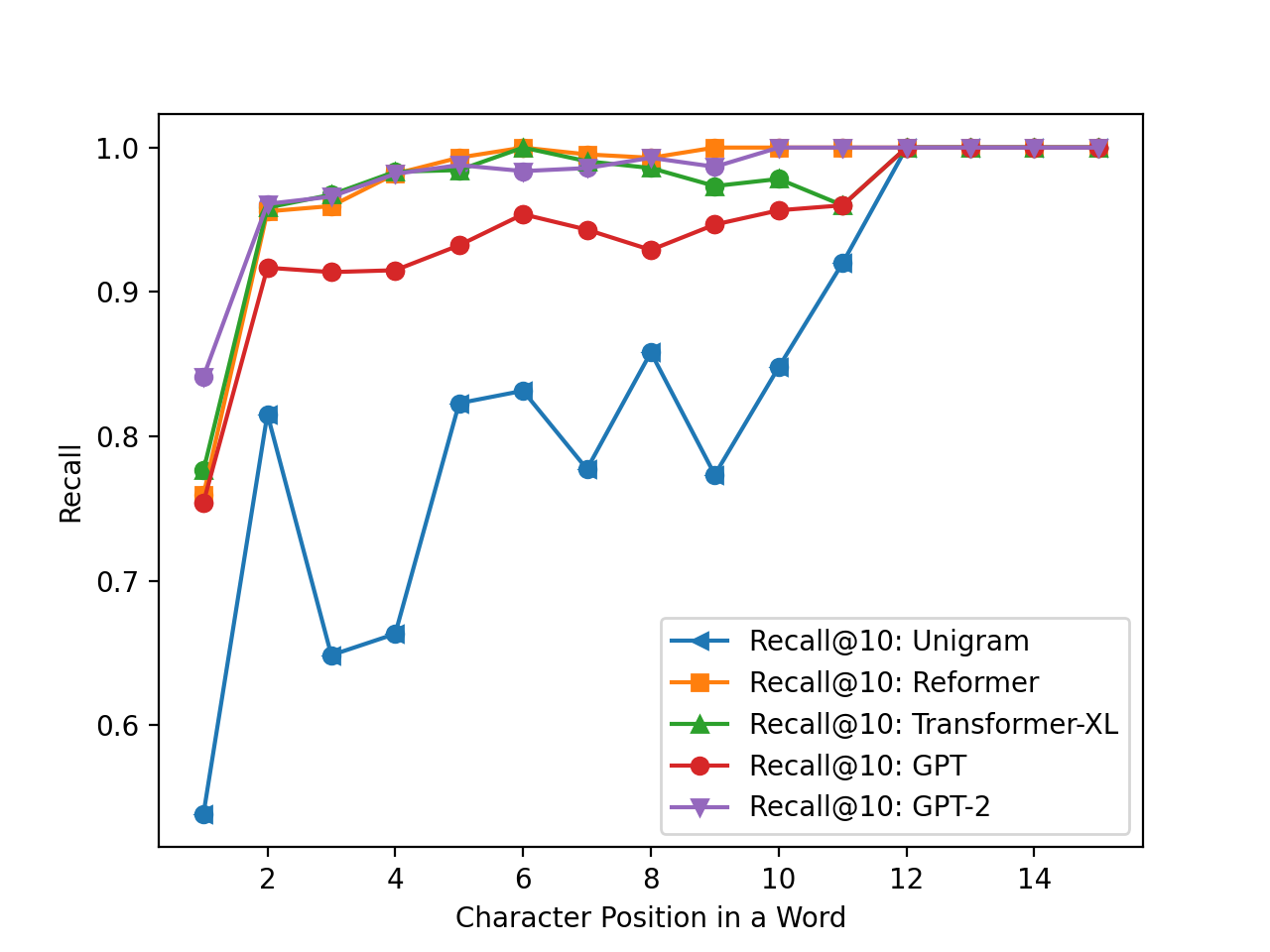}
    \caption{Recall@10 for different character positions on Switchboard Corpus}
    \label{fig:recall@10_swbd_char_pos}
\end{figure}

    Figure~\ref{fig:mrr@10_swbd_char_pos} and Figure \ref{fig:recall@10_swbd_char_pos} show results for different character positions in the predicted words on the Switchboard Corpus. Similar to what we have on the ALS phraseset, predicting the first character in a word is the most difficult for transformer models. When predicting earlier positions in words, there are significant gaps between the results of GPT and that of other transformer models. Such gaps contribute to the fact that GPT performs worst among all transformer models on average over all character positions. The gaps eventually diminish as we have longer context in predicted words. Although both MRR@10 and Recall@10 approach 1 for transformer models towards the end of predicted words, we do see a drop in MRR@10 from GPT and Transformer-XL when the number of context character is 12, which is interesting.

\subsection{The Role of Context Length}
\label{sec: context_length}

In this section we discuss how performance varies with different context lengths in input history. We represent context length as the number of words in the input prior to the words used for prediction. Since the context lengths in the Switchboard corpus are scattered sparsely over a large range of numbers, the results do not provide very meaningful interpretations. Therefore, we only show results on the ALS Phraseset in this section.

\begin{figure}[h]
    \includegraphics[width=\linewidth]{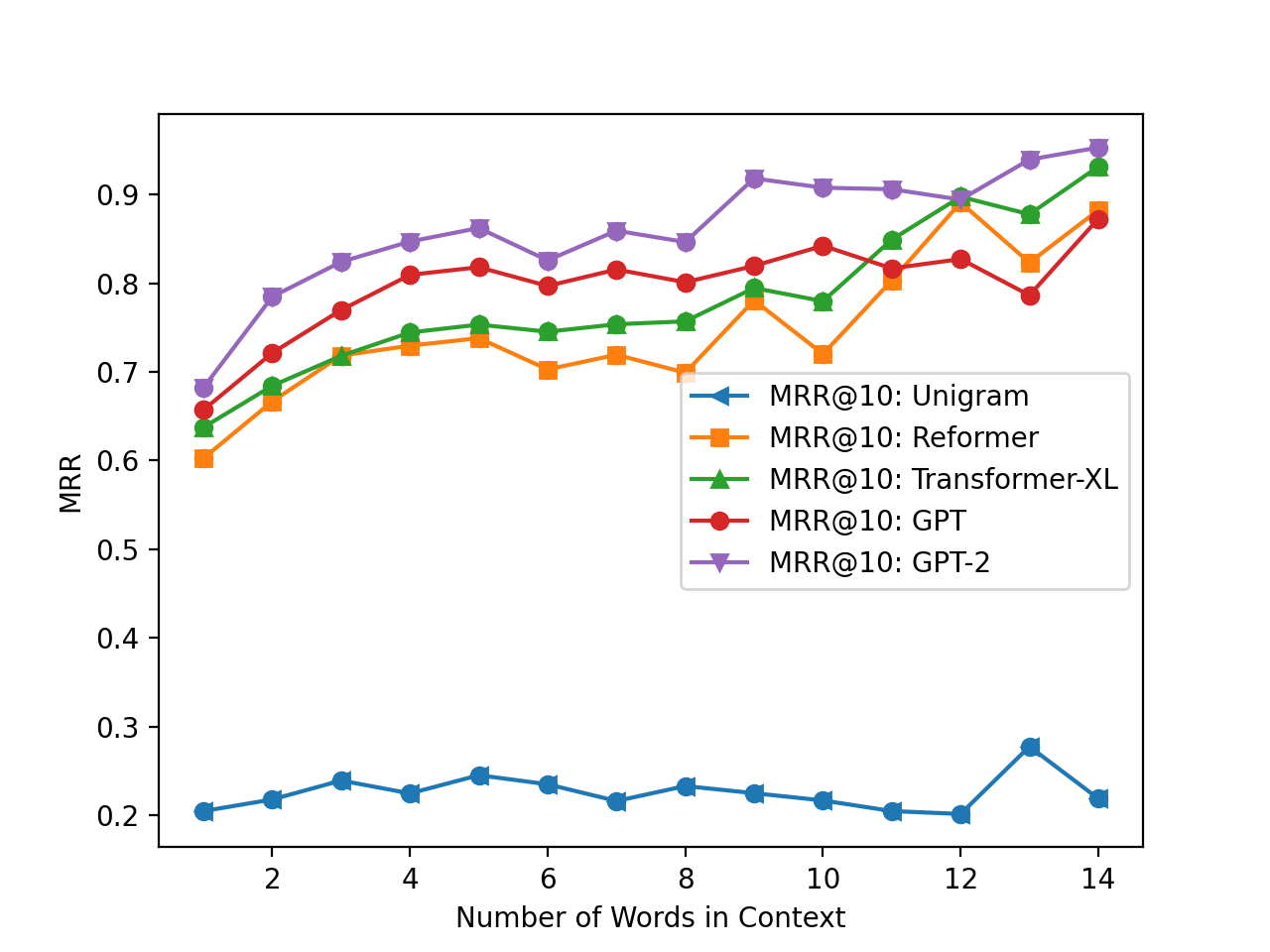}
    \caption{MRR@10 for different number of context words on ALS Phraseset}
    \label{fig:mrr@10_cxt_als}
\end{figure}

\begin{figure}[h]
    \includegraphics[width=\linewidth]{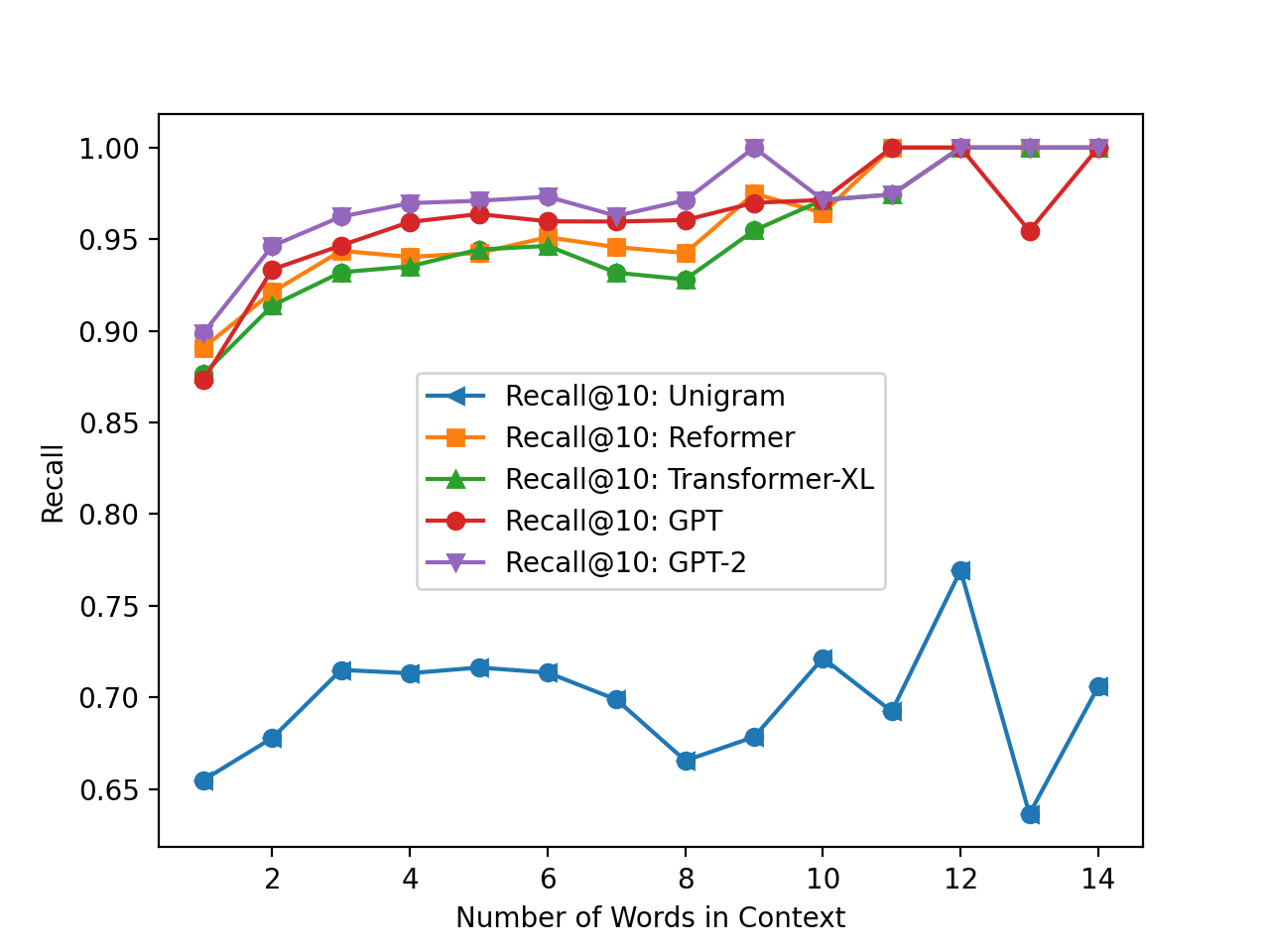}
    \caption{Recall@10 for different number of context words on ALS Phraseset}
    \label{fig:recall@10_cxt_als}
\end{figure}

%\begin{figure}
%    \includegraphics[width=\linewidth]{als_mrr_context_first_char.png}
%    \caption{MRR@10 vs. context length on predicting the first characters of words on ALS Phraseset}
%    \label{fig:mrr@10_cxt_als_first_char}
%\end{figure}

%\begin{figure}
%    \includegraphics[width=\linewidth]{als_recall_context_first_char.png}
%    \caption{Recall@10 vs. context length on predicting the first characters of words on ALS Phraseset}
%    \label{fig:recall@10_cxt_als_first_char}
%\end{figure}

%\begin{figure}
%    \includegraphics[width=\linewidth]{mrr_context_swbd.png}
%    \caption{MRR@10 for different number of context words on Switchboard Corpus}
%    \label{fig:mrr@10_cxt_swbd}
%\end{figure}

%\begin{figure}
%    \includegraphics[width=\linewidth]{recall_context_swbd.png}
%    \caption{Recall@10 for different number of context words on Switchboard Corpus}
%    \label{fig:recall@10_cxt_swbd}
%\end{figure}

\paragraph{Average performance over all character positions} Figure~\ref{fig:mrr@10_cxt_als} and Figure \ref{fig:recall@10_cxt_als} show MRR@10 and Recall@10 averaged over all character positions in predicted words for different number of context words. As we can see, results on both metrics increase as we have more words in context for transformer models. Of all transformer models, GPT-2 gives best results on both MRR@10 and Recall@10 for almost all different context lengths, which is aligned with our observations from \cref{sec:main_results_char_pred} that GPT2 fares best across all models. 
    
    %\item \textbf{Performance on predicting the first character}: Figure \ref{fig:mrr@10_cxt_als_first_char} and Figure \ref{fig:recall@10_cxt_als_first_char} illustrate MRR@10 and recall@10 on predicting the first character of words for different context length. Unlike what we have for the average performance, results on MRR@10 do not show a strictly upward trend as the length of available context increases. However, recall@10 generally has a positive correlation between context length and the metric number. Overall (especially when the context length is shorter), TransformerXL and Reformer do not perform as well as GPT and GPT2 when predicting the first character.
    
\begin{figure}[h]
    \includegraphics[width=\linewidth]{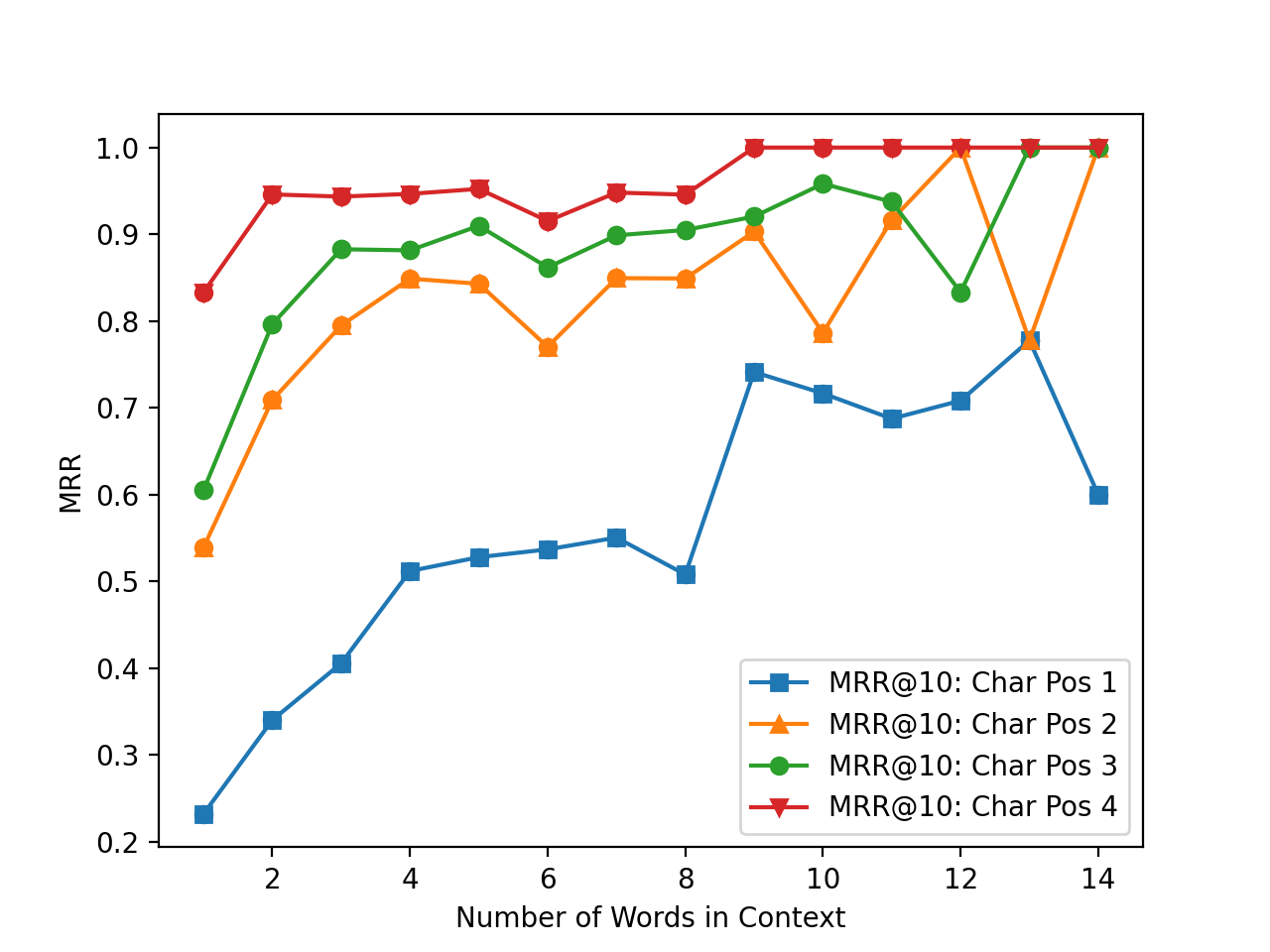}
    \caption{GPT2 model's MRR@10 vs. context length on predicting different character positions of words on ALS Phraseset}
    \label{fig:mrr@10_cxt_als_gpt2_diff_pos}
\end{figure}

\begin{figure}[h]
    \includegraphics[width=\linewidth]{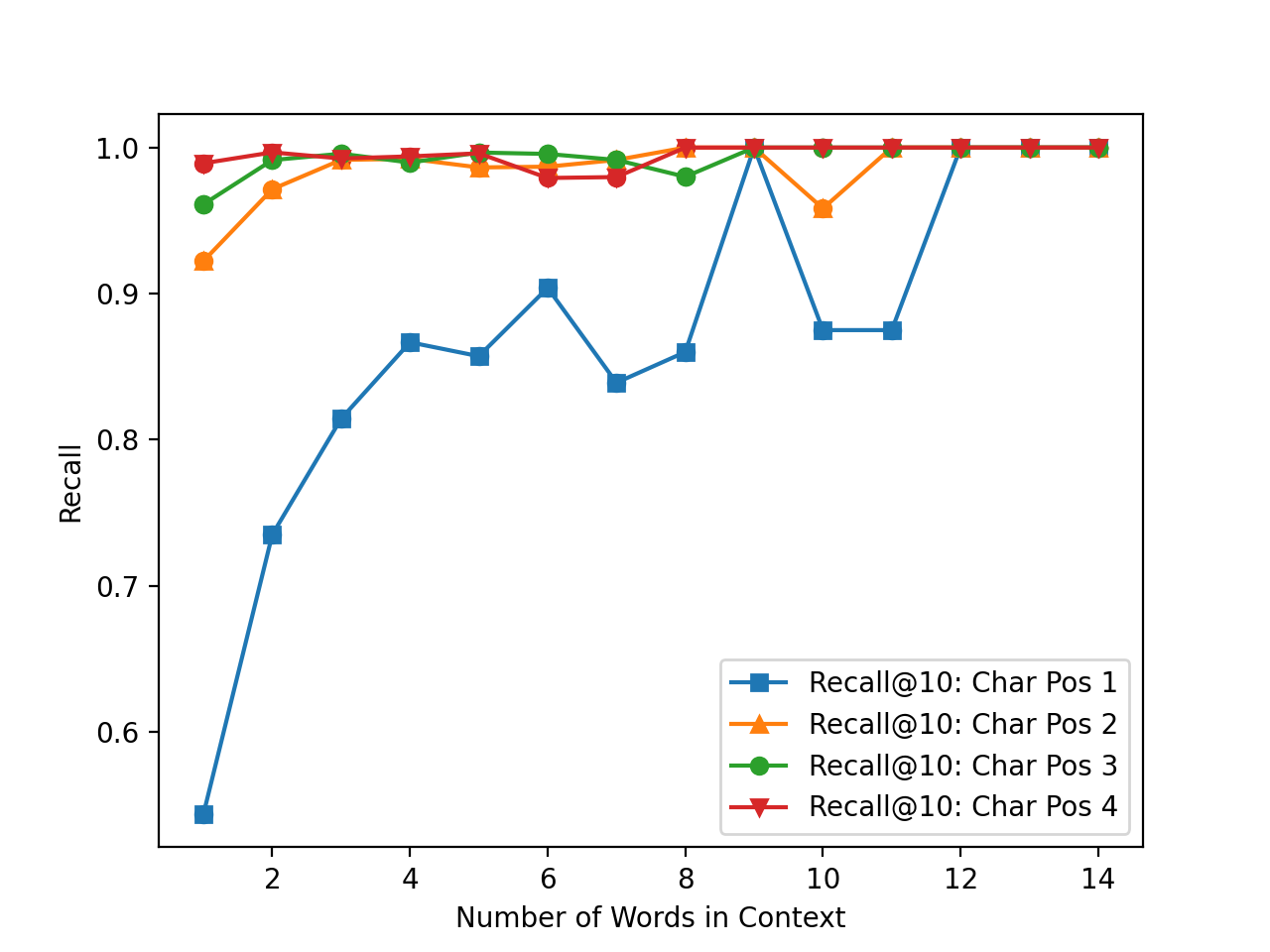}
    \caption{GPT2 model's Recall@10 vs. context length on predicting different character positions of words on ALS Phraseset}
    \label{fig:recall@10_cxt_als_gpt2_diff_pos}
\end{figure}

\paragraph{GPT-2's performance vs. context length on predicting different character positions of words} Figure~\ref{fig:mrr@10_cxt_als_gpt2_diff_pos} and Figure~\ref{fig:recall@10_cxt_als_gpt2_diff_pos} illustrate MRR@10 and Recall@10 on predicting the first, second, third and fourth characters in words versus different number of context words from the GPT-2 model. As shown in the results, the performance on predicting the first character is consistently the worst. The results on both metrics also show a general improvement in performance when the character position increases. There is no strict upward trend in MRR as we have more context, especially for predicting the first character.

%\begin{figure}
%    \includegraphics[width=\linewidth]{gpt2_mrr_diff_pos_cxt_swbd.png}
%    \caption{GPT2 model's MRR@10 vs. context length on predicting different character positions of words on Switchboard Corpus}
%    \label{fig:mrr@10_cxt_swbd_gpt2_diff_pos}
%\end{figure}

%\begin{figure}
%    \includegraphics[width=\linewidth]{gpt2_recall_diff_pos_cxt_swbd.png}
%    \caption{GPT2 model's Recall@10 vs. context length on predicting different character positions of words on Switchboard Corpus}
%    \label{fig:recall@10_cxt_swbd_gpt2_diff_pos}
%\end{figure}

\subsection{Noisy Histories}

In this section, we examine the effect of noise in typing history and how it changes the prediction performance of different models. Noisy input is very common in BCI settings since BCI users often make spelling mistakes. In our experiment, we artificially replace a certain amount of context with randomly selected letters from the English alphabet. % We experiment on the scenarios where spaces cannot and can be substituted in the noisy input. We do not change the spaces in context since we believe that users of BCI systems should be able to identify word boundaries and type spaces correctly. However, they may make mistakes when typing the actual words.

Table \ref{tab:als_phraseset_noise_delta_10_space} and Table \ref{tab:switchboard_noise_delta_10_space} show the decreases in performance metrics for various models on the ALS Phraseset and on the Switchboard Corpus. Absolute numbers of the performance metrics are in appendix \ref{sec: appendix}.
%where Table \ref{tab:als_phraseset_noise_delta_10_space} and Table \ref{tab:switchboard_noise_delta_10_space} (In appendix \ref{sec: appendix}) demonstrates the same metrics where spaces can be substituted (For all cases when set the noise rate to be 10\%). 
On both datasets, Transformer-XL exhibits the smallest degradation in performance among all models. Furthermore, Transformer-XL shows the best absolute performance across all metrics. This is probably due to the fact that, since Transformer-XL employs a closed vocabulary, we do not use beam search to predict multiple subword units. Instead we let the model predict the last word of input and compute exact character marginals on words in the vocabulary that have prefixes matching the partially typed last word. Therefore, the effect of noise is smaller since the model considers longer input units (words instead of subwords or individual characters). When a single incorrect character appears in a relatively long whole word, it may not confuse the model very much. On the other hand, both GPT and GPT-2 see significant drops in performance with the presence of noise, indicating that these models are not very robust to noisy input. 

\begin{figure}
    \includegraphics[width=\linewidth]{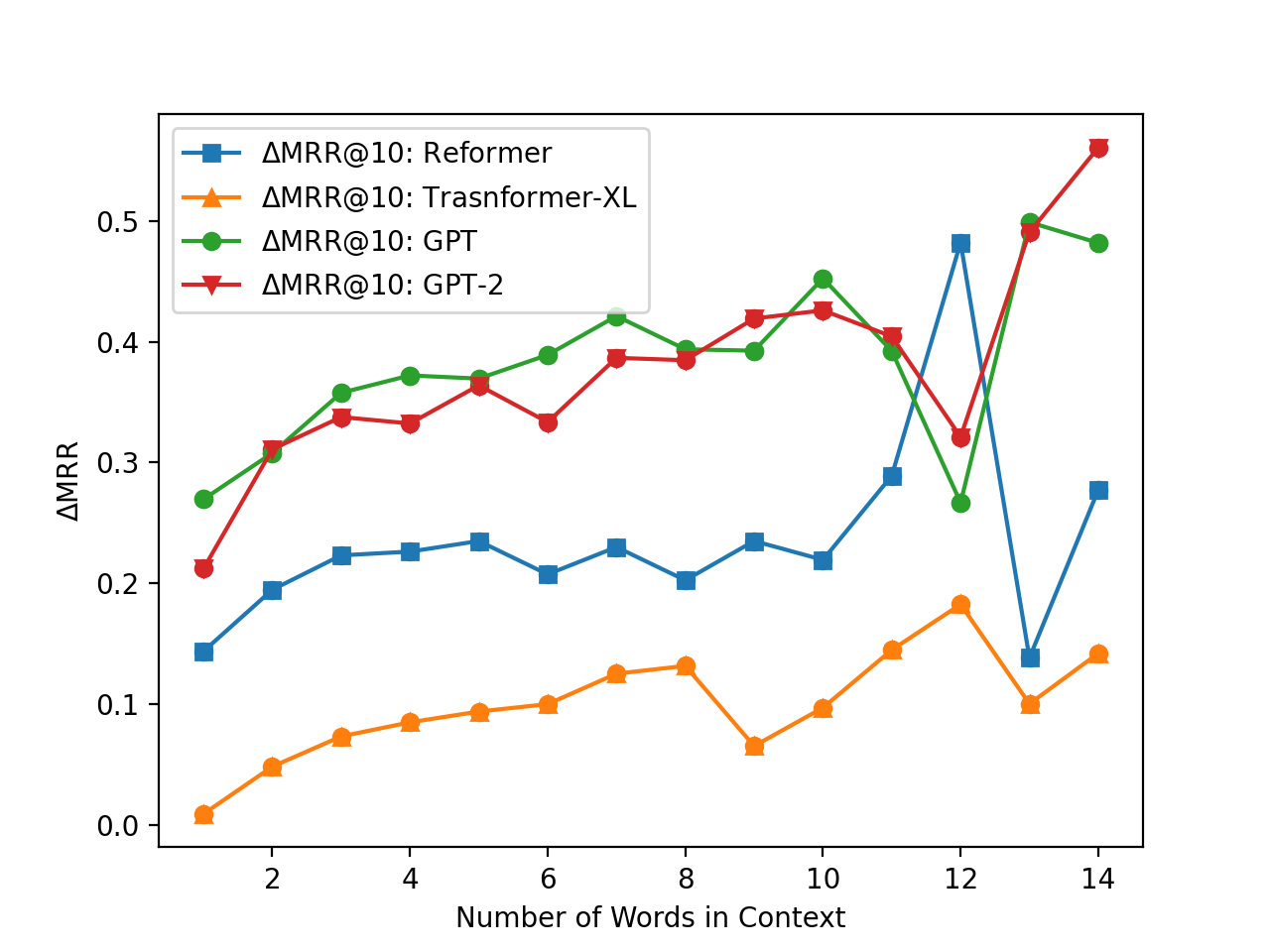}
    \caption{$\Delta$MRR@10 for different number of context words on ALS Phraseset with 10\% noise present}
    \label{fig:mrr@10_cxt_als_10_percent_noise}
\end{figure}

\begin{figure}
    \includegraphics[width=\linewidth]{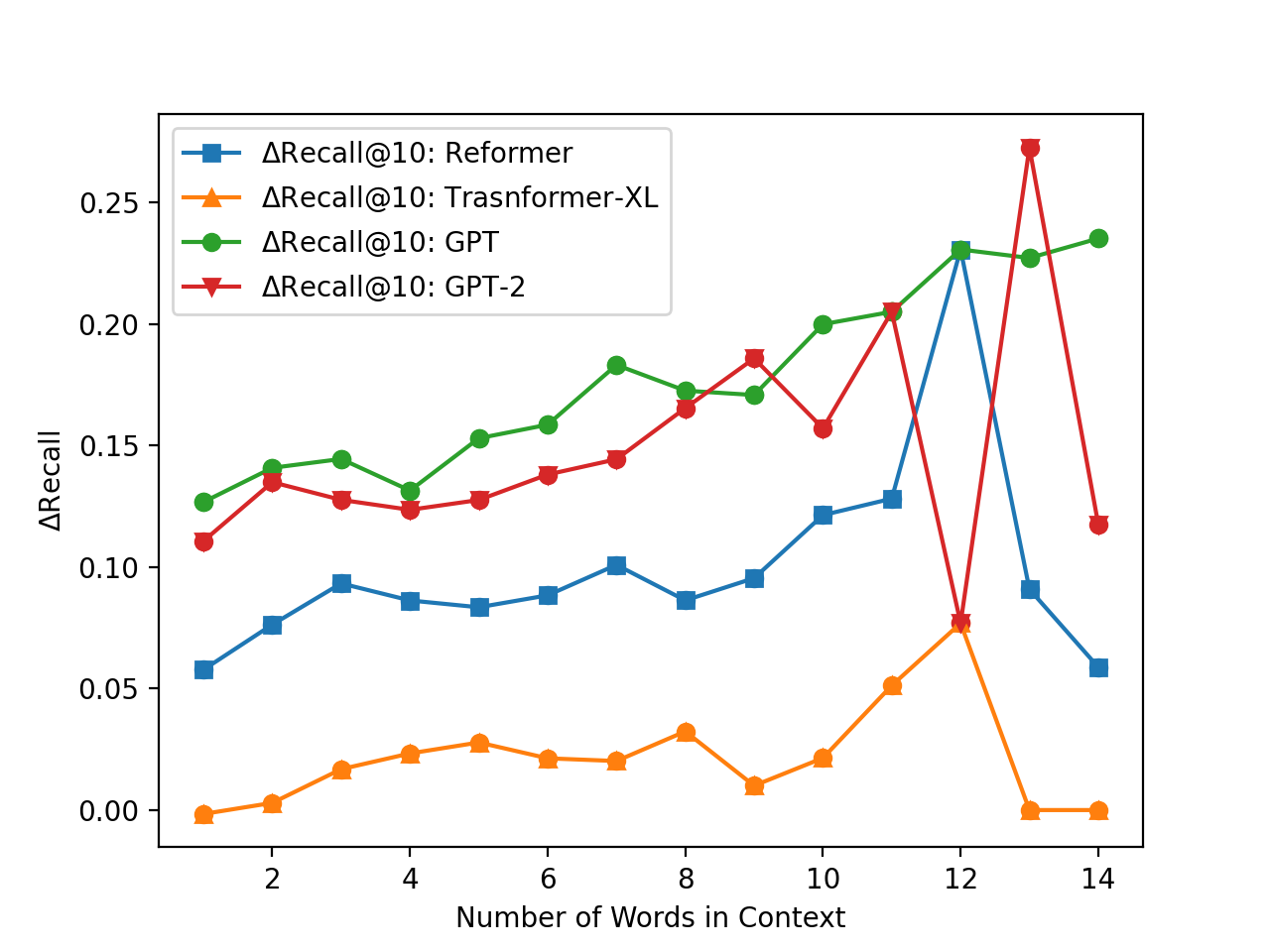}
    \caption{$\Delta$Recall@10 for different number of context words on ALS Phraseset with 10\% noise present}
    \label{fig:recall@10_cxt_als_10_percent_noise}
\end{figure}

Figure \ref{fig:mrr@10_cxt_als_10_percent_noise} and Figure \ref{fig:recall@10_cxt_als_10_percent_noise} illustrate the changes in MRR and Recall for different models with respect to different context lengths (As in Section~\ref{sec: context_length} above, we only experiment on the ALS phraseset). In general, the drops in two metrics increases when the context is longer. This is probably because more noise (in an absolute sense) is introduced with longer input, which brings more negative impact on predictions given by the language models. However, we do see a few outliers where the changes drop sharply with longer context. This is an interesting phenomenon, and we would like to explore more on it for future work.

\begin{table*}
\centering
    \begin{tabular}{c|cccccc}
    \hline
    Language Model & $\Delta$MRR@10 & $\Delta$Recall@10 & $\Delta$MRR@5 & $\Delta$Recall@5 & $\Delta$MRR@3 & $\Delta$Recall@3 \\
    \hline
    Reformer & -0.0903 & -0.0338 & -0.0947 & -0.0662 & -0.0990 & -0.0846\\
    Transformer XL & \textbf{-0.0401} & \textbf{-0.0079} & \textbf{-0.0423} & \textbf{-0.0231} & \textbf{-0.0448} & \textbf{-0.0332} \\
    GPT: Beam Search & -0.2089 & -0.0777 & -0.2192 & -0.1546 & -0.2294 & -0.1999 \\
    GPT2: Beam Search & -0.1863 & -0.0631 & -0.1940 & -0.1201 & -0.2040 & -0.1633 \\
    \hline
    \end{tabular}
    \caption{Results Difference on the ALS Phraseset with 10\% noise present, averaged across all character positions}
    \label{tab:als_phraseset_noise_delta_10_space}
\end{table*}

\begin{table*}
\centering
    \begin{tabular}{c|cccccc}
    \hline
    Language Model & $\Delta$MRR@10 & $\Delta$Recall@10 & $\Delta$MRR@5 & $\Delta$Recall@5 & $\Delta$MRR@3 & $\Delta$Recall@3 \\
    \hline
    Reformer & -0.1162 & -0.0453 & -0.1218 & -0.0891 & -0.1267 & -0.1097 \\
    Transformer XL & \textbf{-0.0693} & \textbf{-0.0245} & \textbf{-0.0702} & \textbf{-0.0319} & \textbf{-0.0753} & \textbf{-0.0532} \\
    GPT: Beam Search & -0.1607 & -0.0713 & -0.1677 & -0.1244 & -0.1756 & -0.1588 \\
    GPT2: Beam Search & -0.1935 & -0.0708 & -0.2013 & -0.1288 & -0.2093 & -0.1643 \\
    \hline
    \end{tabular}
    \caption{Results Difference on the Switchboard Corpus with 10\% noise present, averaged across all character positions}
    \label{tab:switchboard_noise_delta_10_space}
\end{table*}

%\begin{figure}
%    \includegraphics[width=\linewidth]{noise_40_percent_delta_mrr_als_new.png}
%    \caption{$\Delta$MRR@10 for different number of context words on ALS Phraseset with 40\% noise present}
%    \label{fig:mrr@10_cxt_als_40_percent_noise}
%\end{figure}

%\begin{figure}
%    \includegraphics[width=\linewidth]{noise_40_percent_delta_recall_als_new.png}
%    \caption{$\Delta$Recall@10 for different number of context words on ALS Phraseset with 40\% noise present}
%    \label{fig:recall@10_cxt_als_40_percent_noise}
% \end{figure}

%\subsection{Further Experiments}

%\begin{itemize}
   % \item \textbf{Performance with different context length} For both datasets, compare the MRR and recall results with varied context length. The context length can be different number of words and number of characters. Also, for different context lengths, compare the performance on the first character vs. that on subsequent characters. Take a few examples where the prediction of the first character does not fare well and examine why.
   % \item \textbf{MacKenzie and Soukoreff phraseset} \citep{mackenzie2003phrase} Experiments on this dataset as well. This dataset contains a collection of 500 phrases for evaluation of text entry methods.
   % \item \textbf{Optimizing rescaling factor $\alpha$} Should we optimize $\alpha$ for different EEG response? Make a scatter plot of entropy of LM predictions vs. optimal $\alpha$. Optimal $\alpha$ being the largest $\alpha$ that would make the EEG response bump the target letter to top rank given that it is not at the top in the language model prediction. 
%\end{itemize}

\section{Conclusion}

In this paper we explore the performance of four transformer language models (Reformer, Transformer-XL, GPT-2 and GPT) on predictive typing tasks, using metrics relevant to their deployment in brain-computer interface systems. We experiment on two datasets: the ALS Phraseset and the Switchboard corpus. We also develop a beam search based algorithm for GPT-2 and GPT to predict multiple subword units. We examine the experimental results averaged over all character positions in a word, as well as on different character positions. Among all the models, GPT-2 fares best in most scenarios. We also see that the performance improves for all models on later character positions in a word. Furthermore, we investigate the effect of context lengths in prediction performance. Generally, the predictions are more accurate with longer context. Lastly, we examine the effect of noise in input. It turns out that Transformer-XL is the most noise-robust language model at the level of 10\% noise. On the other hand, the performance of GPT-2 and GPT are heavily impacted by the presence of noise. The performance degradation with noise is generally correlated with context lengths.

Future work will involve investigating the relationship between prediction performance and the internal properties of language models. For example, we explore how perplexity relates to these predictive evaluation metrics. Moreover, we can see whether fine-tuning the language models with noisy input could reduce the performance degradation with noisy context. Lastly, we are happy to share that our collaborators have begun testing these language models on real BCI systems. It would be interesting to see how actual users react to the enhancements these language models provide. Overall, we believe that our experiments open exciting avenues for this line of research. 

\clearpage

\section{Limitations}

We have conducted an extensive evaluation of the performance of large language models using the transformer architecture on character-level predictive typing tasks. We note, however, a few areas of limitation that suggest avenues for future work:

\begin{itemize}
    \item As noted above, models using subword tokenization seem particularly susceptible to character typing errors in the history. More work needs to be done exploring how character-level edits change tokenization, how they interact with search errors in beam search, and how these cascading errors can be mitigated.
    \item Due to a lack of extensive, anonymized BCI typing logs, we use a character unigram edit model to simulate noisy data. As more real BCI interaction data becomes available, we could simulate more realistic noisy histories.
    \item Analyzing typing accuracy in longer documents or long-running dialogues, where transformer architectures can take advantage of more context, is left for future work.
\end{itemize}

\clearpage
\bibliography{anthology,custom}

\begin{thebibliography}{20}
\expandafter\ifx\csname natexlab\endcsname\relax\def\natexlab#1{#1}\fi

\bibitem[{Birbaumer et~al.(1999)Birbaumer, Ghanayim, Hinterberger, Iversen,
  Kotchoubey, K{\"u}bler, Perelmouter, Taub, and Flor}]{birbaumer1999spelling}
Niels Birbaumer, Nimr Ghanayim, Thilo Hinterberger, Iver Iversen, Boris
  Kotchoubey, Andrea K{\"u}bler, Juri Perelmouter, Edward Taub, and Herta Flor.
  1999.
\newblock A spelling device for the paralysed.
\newblock \emph{Nature}, 398(6725):297--298.

\bibitem[{Costello(2014)}]{costello2014message}
John~M Costello. 2014.
\newblock Message banking, voice banking and legacy messages.
\newblock \emph{Boston, MA: Boston Children’s Hospital}.

\bibitem[{Dai et~al.(2019)Dai, Yang, Yang, Carbonell, Le, and
  Salakhutdinov}]{dai-etal-2019-transformer}
Zihang Dai, Zhilin Yang, Yiming Yang, Jaime Carbonell, Quoc Le, and Ruslan
  Salakhutdinov. 2019.
\newblock \href {https://doi.org/10.18653/v1/P19-1285} {Transformer-{XL}:
  Attentive language models beyond a fixed-length context}.
\newblock In \emph{Proceedings of the 57th Annual Meeting of the Association
  for Computational Linguistics}, pages 2978--2988, Florence, Italy.
  Association for Computational Linguistics.

\bibitem[{Dong et~al.(2019)Dong, Smith, Dudy, and Bedrick}]{dong2019noisy}
Rui Dong, David~A Smith, Shiran Dudy, and Steven Bedrick. 2019.
\newblock Noisy neural language modeling for typing prediction in bci
  communication.
\newblock In \emph{Proceedings of the Eighth Workshop on Speech and Language
  Processing for Assistive Technologies}, pages 44--51.

\bibitem[{Dudy et~al.(2018)Dudy, Bedrick, Xu, and Smith}]{dudy2018multi}
Shiran Dudy, Steven Bedrick, Shaobin Xu, and David~A Smith. 2018.
\newblock A multi-context character prediction model for a brain-computer
  interface.
\newblock In \emph{Proceedings of the conference. Association for Computational
  Linguistics. North American Chapter. Meeting}, volume 2018, page~72. NIH
  Public Access.

\bibitem[{Ghosh and Kristensson(2017)}]{ghosh2017neural}
Shaona Ghosh and Per~Ola Kristensson. 2017.
\newblock Neural networks for text correction and completion in keyboard
  decoding.
\newblock \emph{arXiv preprint arXiv:1709.06429}.

\bibitem[{Godfrey et~al.(1992)Godfrey, Holliman, and
  McDaniel}]{godfrey1992switchboard}
John~J Godfrey, Edward~C Holliman, and Jane McDaniel. 1992.
\newblock Switchboard: Telephone speech corpus for research and development.
\newblock In \emph{Acoustics, Speech, and Signal Processing, IEEE International
  Conference on}, volume~1, pages 517--520. IEEE Computer Society.

\bibitem[{Hutter(2018)}]{hutter2018}
Marcus Hutter. 2018.
\newblock The human knowledge compression contest.

\bibitem[{Kitaev et~al.(2020)Kitaev, Kaiser, and Levskaya}]{Kitaev2020Reformer}
Nikita Kitaev, Lukasz Kaiser, and Anselm Levskaya. 2020.
\newblock \href {https://openreview.net/forum?id=rkgNKkHtvB} {Reformer: The
  efficient transformer}.
\newblock In \emph{International Conference on Learning Representations}.

\bibitem[{Mora-Cortes et~al.(2014)Mora-Cortes, Manyakov, Chumerin, and
  Van~Hulle}]{mora2014language}
Anderson Mora-Cortes, Nikolay~V Manyakov, Nikolay Chumerin, and Marc~M
  Van~Hulle. 2014.
\newblock Language model applications to spelling with brain-computer
  interfaces.
\newblock \emph{Sensors}, 14(4):5967--5993.

\bibitem[{Oken et~al.(2014)Oken, Orhan, Roark, Erdogmus, Fowler, Mooney,
  Peters, Miller, and Fried-Oken}]{oken2014brain}
Barry~S Oken, Umut Orhan, Brian Roark, Deniz Erdogmus, Andrew Fowler, Aimee
  Mooney, Betts Peters, Meghan Miller, and Melanie~B Fried-Oken. 2014.
\newblock Brain--computer interface with language model--electroencephalography
  fusion for locked-in syndrome.
\newblock \emph{Neurorehabilitation and neural repair}, 28(4):387--394.

\bibitem[{Radford et~al.(2018)Radford, Narasimhan, Salimans, and
  Sutskever}]{radford2018improving}
Alec Radford, Karthik Narasimhan, Tim Salimans, and Ilya Sutskever. 2018.
\newblock Improving language understanding by generative pre-training.

\bibitem[{Radford et~al.(2019)Radford, Wu, Child, Luan, Amodei, Sutskever
  et~al.}]{radford2019language}
Alec Radford, Jeffrey Wu, Rewon Child, David Luan, Dario Amodei, Ilya
  Sutskever, et~al. 2019.
\newblock Language models are unsupervised multitask learners.
\newblock \emph{OpenAI blog}, 1(8):9.

\bibitem[{Sellers et~al.(2010)Sellers, Vaughan, and Wolpaw}]{sellers2010brain}
Eric~W Sellers, Theresa~M Vaughan, and Jonathan~R Wolpaw. 2010.
\newblock A brain-computer interface for long-term independent home use.
\newblock \emph{Amyotrophic lateral sclerosis}, 11(5):449--455.

\bibitem[{Sennrich et~al.(2016)Sennrich, Haddow, and
  Birch}]{sennrich-etal-2016-neural}
Rico Sennrich, Barry Haddow, and Alexandra Birch. 2016.
\newblock \href {https://doi.org/10.18653/v1/P16-1162} {Neural machine
  translation of rare words with subword units}.
\newblock In \emph{Proceedings of the 54th Annual Meeting of the Association
  for Computational Linguistics (Volume 1: Long Papers)}, pages 1715--1725,
  Berlin, Germany. Association for Computational Linguistics.

\bibitem[{Speier et~al.(2016)Speier, Arnold, and
  Pouratian}]{speier2016integrating}
William Speier, C~Arnold, and Nader Pouratian. 2016.
\newblock Integrating language models into classifiers for bci communication: a
  review.
\newblock \emph{Journal of neural engineering}, 13(3):031002.

\bibitem[{Speier et~al.(2018)Speier, Arnold, Chandravadia, Roberts, Pendekanti,
  and Pouratian}]{speier2018improving}
William Speier, Corey Arnold, Nand Chandravadia, Dustin Roberts, Shrita
  Pendekanti, and Nader Pouratian. 2018.
\newblock Improving p300 spelling rate using language models and predictive
  spelling.
\newblock \emph{Brain-Computer Interfaces}, 5(1):13--22.

\bibitem[{Speier et~al.(2017)Speier, Chandravadia, Roberts, Pendekanti, and
  Pouratian}]{speier2017online}
William Speier, Nand Chandravadia, Dustin Roberts, Shrita Pendekanti, and Nader
  Pouratian. 2017.
\newblock Online bci typing using language model classifiers by als patients in
  their homes.
\newblock \emph{Brain-Computer Interfaces}, 4(1-2):114--121.

\bibitem[{Vaswani et~al.(2017)Vaswani, Shazeer, Parmar, Uszkoreit, Jones,
  Gomez, Kaiser, and Polosukhin}]{vaswani2017attention}
Ashish Vaswani, Noam Shazeer, Niki Parmar, Jakob Uszkoreit, Llion Jones,
  Aidan~N Gomez, {\L}ukasz Kaiser, and Illia Polosukhin. 2017.
\newblock Attention is all you need.
\newblock \emph{Advances in neural information processing systems}, 30.

\bibitem[{Zhu et~al.(2015)Zhu, Kiros, Zemel, Salakhutdinov, Urtasun, Torralba,
  and Fidler}]{zhu2015aligning}
Yukun Zhu, Ryan Kiros, Rich Zemel, Ruslan Salakhutdinov, Raquel Urtasun,
  Antonio Torralba, and Sanja Fidler. 2015.
\newblock Aligning books and movies: Towards story-like visual explanations by
  watching movies and reading books.
\newblock In \emph{Proceedings of the IEEE international conference on computer
  vision}, pages 19--27.

\end{thebibliography}
\bibliographystyle{acl_natbib}

\appendix

\section{Further Noisy Input Results}
\label{sec: appendix}

Table \ref{tab:als_phraseset_noise_10_space} and Table \ref{tab:switchboard_noise_10_space} show the absolute values of performance metrics with 10\% noise present. As we can see, Transformer XL delivers the best results among all models on both datasets. Further, all transformer language models perform significantly better than the unigram baseline even with the presence of noise. 

\begin{table*}
\centering
    \begin{tabular}{c|cccccc}
    \hline
    Language Model & MRR@10 & Recall@10 & MRR@5 & Recall@5 & MRR@3 & Recall@3 \\
    \hline
    Unigram Baseline & 0.2294 & 0.7022 & 0.1835 & 0.3681 & 0.1552 & 0.2424 \\
    Reformer & 0.6177 & 0.9043 & 0.6037 & 0.8002 & 0.5816 & 0.7037\\
    Transformer XL & \textbf{0.6866} & \textbf{0.9229} & \textbf{0.6762} & \textbf{0.8469} & \textbf{0.6591} & \textbf{0.7728} \\
    GPT: Beam Search & 0.5687 & 0.8710 & 0.5514 & 0.7427 & 0.5313 & 0.6546 \\
    GPT2: Beam Search & 0.6390 & 0.8980 & 0.6256 & 0.7994 & 0.6063 & 0.7156 \\
    \hline
    \end{tabular}
    \caption{Results on the ALS Phraseset with 10\% noise present, averaged across all character positions}
    \label{tab:als_phraseset_noise_10_space}
\end{table*}

\begin{table*}
\centering
    \begin{tabular}{c|cccccc}
    \hline
    Language Model & MRR@10 & Recall@10 & MRR@5 & Recall@5 & MRR@3 & Recall@3 \\
    \hline
    Unigram Baseline & 0.2292 & 0.7184 & 0.1821 & 0.3669 & 0.1543 & 0.2420 \\
    Reformer & 0.6199 & 0.8941 & 0.6047 & 0.7817 & 0.5848 & 0.6953 \\
    Transformer XL & \textbf{0.6955} & \textbf{0.9174} & \textbf{0.6856} & \textbf{0.8437} & \textbf{0.6674} & \textbf{0.7738} \\
    GPT: Beam Search & 0.4867 & 0.8247 & 0.4639 & 0.6530 & 0.4424 & 0.5587 \\
    GPT2: Beam Search & 0.6250 & 0.8822 & 0.6109 & 0.7767 & 0.5935 & 0.7007 \\
    \hline
    \end{tabular}
    \caption{Results on the Switchboard Corpus with 10\% noise present, averaged across all character positions}
    \label{tab:switchboard_noise_10_space}
\end{table*}

\end{document}